\documentclass[twoside,11pt]{article}

\usepackage{makecell}
\usepackage{jmlr2e}
\usepackage{multirow}

\jmlrheading{1}{2000}{1-48}{4/00}{10/00}{Marina Meil\u{a} and Michael I. Jordan}

\ShortHeadings{Model and Evaluation: Towards Fairness in Multilingual Text Classification}{Lin et al.}
\firstpageno{1}

\begin{document}

\title{Model and Evaluation: Towards Fairness in Multilingual Text Classification}

\author{\name Nankai Lin \email neakail@outlook.com \\
       \addr School of Computer Science and Technology \\
       Guangdong University of Technology \\
       Guangzhou, Guangdong, 510006, China \\
       \AND
       \name Junheng He \email 2539404758@qq.com \\
       \addr School of Computer Science and Technology \\
       Guangdong University of Technology \\
       Guangzhou, Guangdong, 510006, China \\
       \AND
       \name Zhenghang Tang \email 1742717665@qq.com \\
       \addr School of Computer Science and Technology \\
       Guangdong University of Technology \\
       Guangzhou, Guangdong, 510006, China \\
       \AND
       \name Dong Zhou \footnote{Corresponding author} \email dongzhou@gdufs.edu.cn \\
       \addr School of Information Science and Technology \\
       Guangdong University of Foreign Studies \\
       Guangzhou, Guangdong, 510006, China \\
       \AND
       \name Aimin Yang \footnotemark[1] \email amyang@gdut.edu.cn \\
       \addr School of Computer Science and Technology \\
       Guangdong University of Technology \\
       Guangzhou, Guangdong, 510006, China  \\
       $\ast$ Dong Zhou and Aimin Yang are the co-corresponding authors. \\
       }

\editor{Alexander Clark}

\maketitle
\begin{abstract}
Recently, more and more research has focused on addressing bias in text classification models. However, existing research mainly focuses on the fairness of monolingual text classification models, and research on fairness for multilingual text classification is still very limited. In this paper, we focus on the task of multilingual text classification and propose a debiasing framework for multilingual text classification based on contrastive learning. Our proposed method does not rely on any external language resources and can be extended to any other languages. The model contains four modules: multilingual text representation module, language fusion module, text debiasing module, and text classification module. The multilingual text representation module uses a multilingual pre-trained language model to represent the text, the language fusion module makes the semantic spaces of different languages tend to be consistent through contrastive learning, and the text debiasing module uses contrastive learning to make the model unable to identify sensitive attributes’ information. The text classification module completes the basic tasks of multilingual text classification. In addition, the existing research on the fairness of multilingual text classification is relatively simple in the evaluation mode. The evaluation method of fairness is the same as the monolingual equality difference evaluation method, that is, the evaluation is performed on a single language. We propose a multi-dimensional fairness evaluation framework for multilingual text classification, which evaluates the model’s monolingual equality difference, multilingual equality difference, multilingual equality performance difference, and destructiveness of the fairness strategy. We hope that our work can provide a more general debiasing method and a more comprehensive evaluation framework for multilingual text fairness tasks.

\end{abstract}

\section{Introduction}
With the development of artificial intelligence, scholars no longer simply focus on the performance of the model, but also focus on how to solve the social bias existing in the model. The study of detecting and mitigating social bias in artificial intelligence models is not only a matter of social concerns, but also an engineering issue. Natural language processing (NLP) is a representative field where the injection of bias is visible because its training data, the corpus, contains various social concepts that largely influence the inductive learning process of machines. 

In recent years, the issue of fairness in natural language processing has received extensive attention. A large number of studies have shown that there are obvious biases in various tasks of NLP, such as word representation \citep{10.5555/3157382.3157584}, lexical inference \citep{rudinger-etal-2017-social}, coreference resolution \citep{zhao-etal-2018-gender}, text classification \citep{10.1145/3278721.3278729} and sentiment prediction \citep{kiritchenko-mohammad-2018-examining}, etc. 

However, most of the work focuses on natural language processing tasks for single-language debiasing in representation bias, and multi-language natural language processing tasks are still in the exploratory stage. The debiasing task of multilingual models is more challenging than that of single language models. \cite{Costa-juss_Escolano_Basta_Ferrando_Batlle_Kharitonova_2022} found that language-specific pre-trained models exhibited less gender bias than multilingual pre-trained models. In addition, although some datasets for multilingual fairness research have been proposed, these works actually still focus on monolingual language debiasing and evaluation. The mutual influence between languages will lead to a potential bias between the models, and this bias cannot be well resolved by the single language debiasing model.

When dealing with tasks in multiple languages, multilingual models are easier to use than monolingual models because only one model needs to be trained instead of multiple models. However, since the model needs to learn multiple languages, the learning difficulty of the model is increased, and the performance of the model is also affected to a certain extent. Recent studies \citep{wu-dredze-2020-languages} have shown that multilingual text representations do not learn equally high-quality representations for all languages. \cite{wan2022fairness} also pointed out that in multilingual text processing tasks, morphological complexity will cause performance deviations between different languages. Representational units of finer granularity were shown to help eliminate performance disparity though at the cost of longer sequence length, which can have a negative impact on robustness. Therefore, when dealing with multilingual fairness tasks, the fairness of performance between different languages should also be concerned.

Therefore, this paper focuses on the task of multilingual text classification and proposes a framework for debiasing multilingual text classification based on contrastive learning and a multi-dimensional fairness evaluation framework for multilingual text classification. The multilingual text classification debiasing framework consists of four modules: multilingual text representation module, language fusion module, text debiasing module, and text classification module. The multilingual representation module uses a multilingual pre-trained language model to represent the text, the language fusion module makes the semantic spaces of different languages tend to be consistent through contrastive learning. The text debiasing module uses the idea of contrastive learning to shorten the distance between samples with different sensitive attribute values under the same target label, so as to realize the confusion of sensitive attributes in the semantic space, so that the model cannot recognize the information of sensitive attributes. The text classification module completes the basic tasks of multilingual text classification. For the multilingual text classification task, we hope that the evaluation of the model is not limited to the monolingual equality difference evaluation of each language. Therefore, we propose a multi-dimensional multilingual text classification fairness evaluation framework, which respectively evaluates the model’s fairness across individual languages, fairness across all languages, fairness across multilingual performance, and destructiveness of the fairness strategy. We hope that our work can provide a more general debiasing method and a more comprehensive evaluation framework for multilingual text fairness tasks.

\section{Related Work}

\subsection{Monolingual Text Classification and Fairness Research}

With the development of social media, the task of automatically and efficiently classifying text through natural language processing has received increasing attention \citep{yin2021towards,KOCON2021102643,GarcaDaz2022EvaluatingFC}. Some recent studies have shown that bias can affect the performance of the model in this task. \cite{huang-etal-2020-multilingual} argued that whether a statement is considered hate speech depends largely on who the speaker is. \cite{elsafoury-2022-darkness} investigated the causal effect of the social and intersectional bias on the performance and unfairness of hate speech detection models. Therefore, some debiasing methods for this task have also been proposed. \cite{cheng-etal-2021-mitigating} proposed a context-aware and model-agnostic debiasing strategy that leverages a reinforcement learning technique. \cite{han-etal-2021-decoupling} used adversarial learning for debiasing. \cite{ruder-etal-2022-square} pointed out that measuring the model from only one dimension is very limited, for example, only the performance dimension is considered without considering the fairness dimension. Although most of the previous work has considered the performance and fairness dimensions, a lot of debiasing work is carried out for English, and a little work considers the combination of fairness and multilingual tasks.

\subsection{Multilingual Text Classification Research}
Research on multilingualism has proliferated in order to enable models to be applied to different languages and to gain new knowledge from different languages. \cite{mutuvi-etal-2020-multilingual} proposed that text classification models tend to perform differently across different languages, more particularly when the data is highly imbalanced in different languages in the dataset. To solve the performance deviation of different languages, there are three main solution strategies: (1) \textbf{Construct more balanced multilingual datasets}: For example, \cite{ponti-etal-2020-xcopa} proposed a dataset for causal commonsense reasoning in 11 languages XCOPA, \cite{zhao-etal-2020-gender} proposed a multilingual dataset for bias analysis. (2) \textbf{Use multilingual pre-trained language models}: \cite{wu-dredze-2019-beto} explored the performance of mBERT in performing five NLP tasks on 15 languages and found that it effectively learned good multilingual representations. (3) \textbf{Cross-language transfer techniques}: Cross-language techniques migrate modeling approaches that perform well in high resource languages to low resource languages. \cite{DBLP:journals/corr/abs-2006-08881} used a cross-lingual pivoting technique to improve the performance of a multilingual machine translation system. On the other hand, \cite{nooralahzadeh-etal-2020-zero} used meta-learning for cross-lingual text classification model migration, etc. 

\subsection{Contrastive Learning}
As an emerging technology, contrastive learning has been proven to be significantly effective in a number of areas. In recent years, there have also been a number of approaches that use contrastive learning to handle multilingual tasks. \cite{pan-etal-2021-contrastive} proposed that mRASP2 based on contrastive learning improves multilingual machine translation performance by closing the gap among representations of different languages. \cite{wang-etal-2021-contrastive} proposed a multilingual text summarization system CALMS based on contrastive learning strategy. \cite{kumar-etal-2022-mucot}introduced contrastive loss to obtain performance improvements when migrating English-language Question Answering systems to other languages. \cite{pan-etal-2021-contrastive} proposed a multilingual contextual embeddings alignment method based on contrastive learning, which enables mBERT with better results and fewer parameters.
What's more, contrastive learning has also recently been proposed to be applied to fairness. \cite{cheng2020fairfil} proposed a sentence-level debiasing method based on a contrastive learning framework for pre-trained language models, which minimizes the correlation between biased words and preserves rich semantic information of the original sentences. \cite{shen2021contrastive} used contrastive learning to mix text representations with different protected attributes to reduce the correlation between the primary task and the protected attributes. \cite{Gupta} used contrastive learning to limit the mutual information between text representations and protected attributes and reduce the impact of protected attributes on text representation. \cite{ijcai2021p473} proposed a graph debiased contrastive learning framework to optimize the graph representations by aligning with clustered class information, while the optimized graph representations in turn can improve the effectiveness of the clustering task.

\subsection{Research of Multilingual Text Debiasing Methods}
The presence of bias in multilingual text classification tasks is more complex than in monolingual ones, and therefore some studies on multilingual text debiasing have recently emerged. \cite{huang-etal-2020-multilingual} measured four biases on their proposed multilingual Twitter dataset and found that the degree of bias was inconsistent across languages, implying that monolingual debiasing methods are not necessarily effective in another language. \cite{zhao-etal-2020-gender} found that when a model trained on one language is deployed to another language, it also carries bais from the source to the target languages. \cite{wang-etal-2022-assessing} considered languages as fair objects and found that multimodal models are biased against different languages. \cite{10.1145/3442188.3445907} examined the issue of gender bias in translation systems in German, Korean, Portuguese, and Tagalog and found that scaling up language resources may amplify the bias cross-linguistically. \cite{Costa-juss_Escolano_Basta_Ferrando_Batlle_Kharitonova_2022} found that pre-trained models using specific languages exhibited less gender bias than multilingual pre-trained models and that language-specific models were more distracted from gender coding, which helped a lot in debiasing.

\subsection{Research of Multilingual Text Fairness Assessment Methods}
Because equity is a complex and abstract issue, equity measures vary across domains for different tasks, and the fairness measured by each indicator varies. For the text classification task, \cite{10.1145/2090236.2090255} proposed a framework for fair evaluation that guarantees statistical parity, treating people with similar conditions as similarly as possible. For the NER task, \cite{prabhakaran-etal-2019-perturbation} proposed a general evaluation framework and employed it to NRE-related sentiment NLP models and toxicity NLP models by perturbation sensitivity analysis to detect unintended bias present in them. 

The above approaches are more about assessing the fairness of monolingual models, however, multilingualism is more complex compared to monolingualism, and therefore its evaluation metrics will be different, for example, \cite{camara-etal-2022-mapping} suggested that there is limited work on fairness for multilingualism with tasks, and they proposed a new statistical framework to measure four biases in emotion regression tasks for models trained through English, Spanish, and Arabic. \cite{wan2022fairness} proposed an evaluation framework consisting of a 6-layer Transformer to investigate whether the performance differences of different conditional language models are due to inequities across languages. Although these works have all evaluated the fairness of multilingual models, the evaluation perspective still remains in the independent evaluation of each language

As mentioned above, while there have been many approaches to assessing and mitigating bias, most have been conducted in resource-rich monolingual languages such as English, Spanish, etc., and it has been shown that ignoring equity studies in resource-poor languages is itself an inequitable treatment. What works well in one language may not necessarily work in other languages, and different languages can have different levels of bias, etc. Therefore, the study of multilingual equity is an issue that needs attention. Although some datasets for multilingual fairness research have been proposed, these works actually still focus on monolingual language debiasing and evaluation. The mutual influence between languages will lead to a potential bias between the models, and this bias cannot be well resolved by the single language debiasing model. Therefore, this paper studies the fairness of multilingual text classification tasks from the perspectives of model and evaluation to solve the deficiencies in multilingual fairness tasks.

\section{A Debiasing Framework for Multilingual Text Classification Based on Contrastive Learning}
\begin{figure}
  \centering
  \includegraphics[width=1\textwidth]{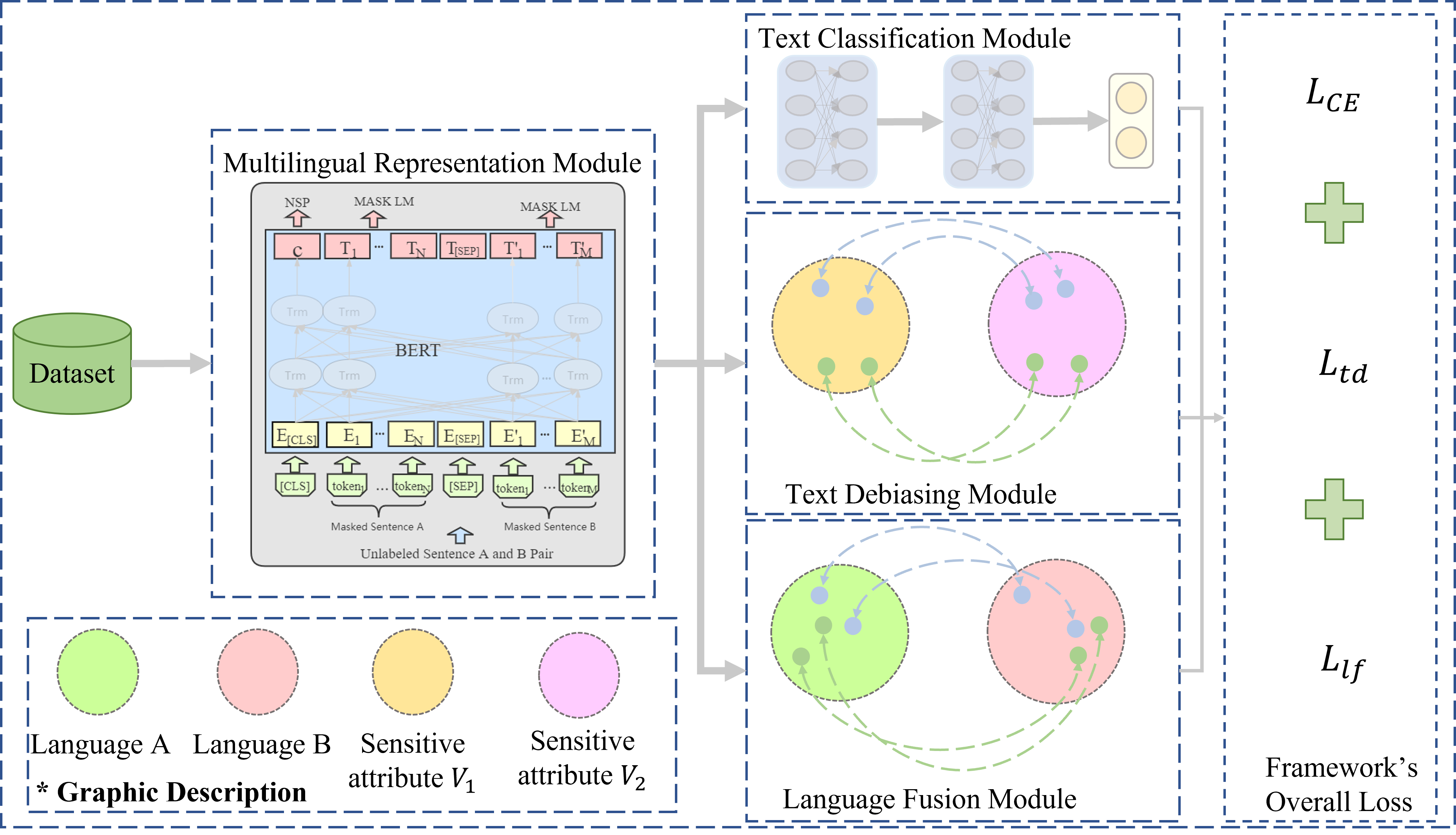}
  \caption{The multilingual text debiasing framework.} 
  \label{fig:1} 
\end{figure}
Figure 1 shows the multilingual text debiasing framework proposed in this paper. The framework we propose can not only guarantee the original classification performance of the model but also achieve an effective debiasing effect. The framework consists of four modules: multilingual representation module, language fusion module, text debiasing module and text classification module. The multilingual representation module uses a multilingual pre-trained language model to represent text, so that data in different languages can be represented in the same encoding method. By using contrastive learning, the language fusion module reduces the distance between samples with the same target label in different languages. As a result, the semantic spaces of different languages tend to be consistent, and the training data in various languages continues to improve the model's overall performance. The text debiasing module employs the concept of contrastive learning to reduce the distance between samples with various sensitive attribute values under the same target label, thereby realizing the confusion of sensitive attributes in the semantic space and preventing the model from recognizing the sensitive attribute information. The text classification module completes the basic tasks of multilingual text classification. 

Given a sample set $G= \{(W_1,y_1,s_1,l_1 ),(W_2,y_2,s_2,l_2 ),…,(W_K,y_K,s_K,l_K)\}$ and $I = \{1,...,K\}$ is the indexes of samples, where $W_i=\{w_1,w_2,…,w_n\} (1 \leq i \leq K)$ is the text sequence of sample $i$, and $n$ is the sentence length of sample $i$. $y_i$ is the target label of sample $i$, where $y_i \in Y$. $Y$ is the category set of the text classification task. $s_i$ is the sensitive attribute that needs to be debiased, where $s_i \in S$, and $S$ is the set of sensitive attribute values. Taking gender as an example, the set $S$ contains two sensitive attribute values, namely male and female. $l_i$ is the language of sample $i$,  where $l_i \in L$, and $L$ is all languages included in the multilingual text classification task. It is worth noting that under the multilingual text task definition, during the training process, the model will receive sensitive attributes and language attributes for training, but during the test process, the model will not treat sensitive attributes and language attributes as known information. 

\subsection{Multilingual text presentation module}
The multilingual representation module aims to represent texts in different languages with the same encoding method, so that the semantic representations of multiple languages are distributed in the same semantic space. We use the mBERT (Multilingual Bidirectional Encoder Representations from Transformers) pre-trained model to encode and represent text. The BERT model is a pre-training model based on Transformer. By using NSP (Next Sentence Prediction) task and MLM (Masked Language Modeling) task for pre-training, the model has rich semantic representation. mBERT is the multilingual text of the BERT model. After the text is segmented using wordpiece, the token, segment and position embeddings of the text are extracted. Besides, generally, we use a special token [CLS] as its sentence vector representation. Therefore, our multilingual representation module utilizes the mBERT model as the base model to encode language features, and uses the first input token [CLS] to obtain sentence vectors express. Therefore, for the $i$-th sentence $W_i$, the sentence vector representation is as follows: 
\begin{equation}
    v_i=mBERT(a_i,b_i,c_i)
\end{equation}

where $a_i$, $b_i$, $c_i$ are the token embeddings, the segmentation embeddings and the position embeddings respectively.

\subsection{Language Fusion Module}

\begin{figure}
  \centering
  \includegraphics[width=1\textwidth]{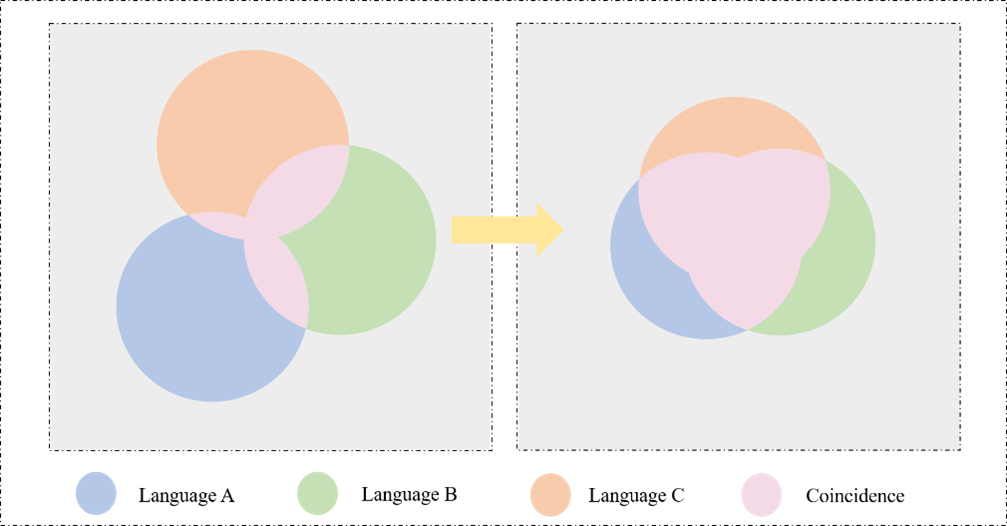}
  \caption{Example of language fusion module effect.} 
  \label{fig:2} 
\end{figure}

The language fusion module aims to make the representation of different languages in the semantic space more compact, as shown in Figure 2, it can be seen that we hope to increase the overlap of the semantic spaces of multiple languages through the language fusion module (that is, in Figure 2 the larger the pink area). In the language fusion module, we narrow the distance between samples with the same target label in different languages through comparative learning, so that the semantic spaces of different languages tend to be consistent. For a given sample i, all other samples that share the same target label with it in the batch form the set $T=\{t:t \in I, y_t=y_i l_t \neq l_i \wedge t \neq i\}$. Among them, the sample set $T$ contains samples of different languages. Then we could define the contrastive loss function of the language fusion module for each entry $i$ across the batch is
\begin{equation}
L_{lf_i}= - \sum_{t \in T } \log \frac
    {\exp({sim(v_i,v_t)} / \tau)}
    { \sum_{k \in I / \{i\}} \exp({sim(v_i,v_k)} / \tau) }
\end{equation}

where $sim(\cdot)$ indicates the cosine similarity function.

The loss of the language fusion module for the entire batch is

\begin{equation}
    L_{lf}= \frac{1}{|G|} \sum_{i=1}^{|G|} L_{lf_i}
\end{equation}

\subsection{Text Debiasing Module}
Unlike other debiasing methods that rely on external resources such as sensitive attribute dictionaries, the text debiasing module can achieve a more general debiasing operation through contrastive learning methods. The language fusion module shortens the semantic distance of samples of the same label in different languages, while the text debiasing module shortens the semantic distance of samples of the same label with different sensitive attribute values.

The text debiasing module aims to make the representation between samples of different sensitive attribute values more closely in the semantic space, so that the model is confused when identifying sensitive attributes, that is, we hope that different sensitive attribute values cannot be distinguished. So for a given sample i, all other samples that share the different sensitive attribute with it in the batch form the set $Q=\{q:q \in I, y_q = y_i \wedge S_q \neq S_i\wedge q \neq i\}$. Then we could define the contrastive loss function of the text debiasing module for each entry i across the batch is
\begin{equation}
L_{td_i}= - \sum_{q \in Q } \log \frac
    {\exp({sim(v_i,v_q)} / \tau)}
    { \sum_{k \in I / \{i\}} \exp({sim(v_i,v_k)} / \tau) }
\end{equation}

Under the entire batch, the loss value obtained by the text debiasing module is
\begin{equation}
    L_{td}= \frac{1}{|G|} \sum_{i=1}^{|G|} L_{td_i}
\end{equation}

\subsection{Text Classification Module}
In the text classification module, we feed the output of the multilingual text presentation module as an input to the feed forward layer with $K × H$ dimensional weight, denoted by $W$, where $K$ is the number of labels. We denote this intermediate representation as F with dimension $1 × K$.
\begin{equation}
    F=v_i \cdot W^T+b
\end{equation}

where $ \cdot $ denotes the dot product between the weight matrix W and the context vector $v_i$ and $b$ is a bias term. The final label probabilities are computed with a standard softmax function using the intermediate representation. The output of the softmax layer P has a dimension of $1 × K$.
\begin{equation}
    P=softmax(F)
\end{equation}

The text classification module uses cross-entropy loss as the training target:

\begin{equation}
 L_{ce}= - \frac{1}{\left| Y \right|} \sum_{n=1}^{\left| Y \right|} y_n \log(P_n)
\end{equation}

\subsection{Framework’s Overall Loss}
The overall training loss of the framework is the overall loss of the language fusion module, text debiasing module, and text classification module:
\begin{equation}
    L= \alpha \cdot L_{lf} + \beta \cdot L_{td} + (1-\alpha-\beta) \cdot L_{ce}
\end{equation}
Among them, $\alpha$ and $\beta$ are the loss weights, which are used to balance the learning intensity of each module. We employ Hyperopt \citep{bergstra2011algorithms} to search for two loss weights.

\section{A Fairness Evaluation Framework for Multilingual Text Classification}
For the multilingual text classification task, we hope that the evaluation of the model is not limited to the monolingual equality difference evaluation of each language. Therefore, we propose a multi-dimensional multilingual text classification fairness evaluation framework, which respectively evaluates the model’s fairness across individual languages, fairness across all languages, fairness across multilingual performance, and destructiveness of the fairness strategy.

\subsection{Monolingual Equality Difference}
As with other existing multilingual fairness studies \cite{10.1145/3278721.3278729,huang-etal-2020-multilingual}, we measure monolingual equality difference (MED) for the multilingual model.

We use the classic monolingual text classification fairness evaluation metric FPED as the evaluation metric of monolingual equality difference. FPED sums the difference between the false positive rate (FPR) within a specific sensitive attribute's group and the false positive rate for all data. We calculate this metric by the following formula:
\begin{equation}
    MED = \sum_{d \in D, l_i \in L} |FPR_{(l_i,d)}-FPR_{l_i}|
\end{equation}

where $D$ is a sensitive attribute (e.g., race) and $d$ is a sensitive attribute group (e.g., white or nonwhite)

\subsection{Multilingual Equality Difference}
Different from multilingual fairness evaluation, multilingual fairness focuses on evaluating the fairness of the model in all languages, that is, when calculating FPR, it is the result of statistics on the test set obtained by combining all languages, that is
\begin{equation}
    MUED= \sum_{d \in D} |FPR_{(L,d)}-FPR_{L}|
\end{equation}

Among them, $FPR_{(L,d)}$ is the false positive rate calculated by the test set samples with the label d of the sensitive attribute D in all languages, and $FPR_{L}$ is the false positive rate calculated by the test set samples in all languages.

\subsection{Multilingual Equality Performance Difference}
Multilingual group fairness not only requires multilingual models to be fair on sensitive attributes, but also wants to achieve equal predictive performance between different languages. From a presentation perspective, it is difficult to implement such a requirement without clearly defined tasks and metrics. We represent multilingual equality performance difference by equalizing the macro-averaged $F$-score across languages. Given a text classification model, its macro-average F value on the test set in different languages is

\begin{equation}
    F= \{ F_{l_1},F_{l_2}…,F_{l_{|L|}} \}
\end{equation}

Next, we take language as group membership and define multilingual equality performance difference by equalizing accuracy across languages. A multilingual model $M$ satisfies multilingual $F$ value parity if $F_{l_i} = F_{l_j}$ for all languages $l_i$, $l_j$. In practice, it is impossible to achieve $F$ value parity for all languages. We first compute the mean of the macro-average F-scores for all languages:

\begin{equation}
    F_{avg} =  \frac{1}{|L|} \sum_{i=1}^{|L|} F_{l_i} 
\end{equation}

Then, similar to how single-language fairness is calculated, we sum the difference between the performance of a specific language and the performance of all languages, which we calculate by the following formula

\begin{equation}
    MEPD= \sum_{l_i \in L} |F_{l_i} - F_{avg}|
\end{equation}
\subsection{Strategy Destructiveness}

We further evaluate whether the model will increase the bias of a sensitive attribute when removing the bias of another sensitive attribute. We hope that the debiasing of the model will not only focus on a certain sensitive attribute to be debiased, but can focus on all sensitive attributes. We believe that when a model debiases one sensitive attribute, it should not negatively affect other sensitive attributes. Take the model debiasing gender sensitive attribute as an example. After debiasing the model, we need to evaluate it on other sensitive attributes such as race and age. A model should not be considered "fair" if it achieves debiasing for gender attributes but adds bias for other attributes like race or age.

Given that the current sensitive attribute of debiasing is $S$, and other existing sensitive attributes are $O={O_1,…,O_z}$, assuming that the fairness results of the model without debiasing operation under other sensitive attributes $O$ are $MED_O$, respectively. The fairness results of the model obtained after debiasing the attribute $S$ under other sensitive attributes $O$ are $MED_{O_{new}}$, and then we define the destructiveness of the debiasing operation on other sensitive attributes as:
\begin{equation}
    SD = \frac{1}{\mid O \mid }  \sum_{i=1}^{\mid O \mid} min(MED_{O_{i,new}} - MED_{O_i},0)
\end{equation}
\section{Experiments and Analysis}

\subsection{Dataset}
We used the multilingual hate speech classification proposed by Huang et al. They collected the tweets annotated by hate speech labels and their corresponding user profiles in English, Italian, Polish, Portuguese and Spanish. The dataset binarizes all tweets’ labels (indicating whether a tweet has indications of hate speech), allowing to merge the different label sets and reduce the data sparsity. Whether a tweet is considered hate speech heavily depends on who the speaker is; for example, whether a racial slur is intended as hate speech depends in part on the speaker’s race \citep{waseem-hovy-2016-hateful}. Therefore, hate speech classifiers may not generalize well across all groups of people, and disparities in the detection of offensive speech could lead to bias in content moderation \citep{Shen_Yoder_Jo_Rose_2018}. We show the corpus statistics in Table 1. Table 1 presents different patterns of the corpus. The Polish data has the smallest number of users. This is because the data focuses on the people who own the most popular accounts in the Polish data \citep{publ152265}, the other data collected tweets randomly. And the dataset shows a much more sparse distribution of the hate speech label than the other languages. The dataset maps the racial outputs into four categories: Asian, Black, Latino and White. For the race and age attributes, we experiment and evaluate with binarization of race and age with roughly balanced distributions (white and nonwhite,  $\leq$ median vs. elder age). Similarly, we binarize the country to indicate if a user is in the main country or not. For example, the majority of users in the English are from the United States (US), therefore, we can binarize the country attributes to indicate if the users are in the US or not.

\begin{table}
\label{tab:accents1}
\caption{Statistical summary of multilingual corpora across English, Italian, Polish, Portuguese and Spanish.}
\centering
\begin{tabular}{ccccc}
\hline
Language & Users & Docs & Tokens & HS Ratio \\
\hline
English & 64,067 & 83,077 & 20.066 & 0.37 \\
Italian & 3,810 & 5,671 & 19.721 & 0.195 \\
Polish & 86 & 10,919 & 14.285 & 0.089 \\
Portuguese & 600 & 1,852 & 18.494 & 0.205 \\
Spanish & 4,600 & 4,831 & 19.199 & 0.397 \\
\hline
\end{tabular}
\end{table}

\subsection{Experimental Settings}
All experiments were carried out using PyTorch\footnote{https://github.com/pytorch/pytorch} and an RTX 8000 with 48 GB of memory. We build our framework based on Transformers\footnote{https://github.com/huggingface/transformers}. Furthermore, we choose the bert-base-multilingual-uncased\footnote{https://huggingface.co/bert-base-multilingual-uncased} \citep{devlin-etal-2019-bert} model as the pre-trained model adopted by our framework. We set the same hyper-parameters with a fixed initialization seed for our models training, where the batch size is 32 and the feature dimension is 768. The num of epoch is 10 and the max length of the input samples is 32. With a learning rate of 5e-5 for the BERT encoder, Adam was chosen for optimization. For the loss weight $\alpha$, $\beta$ and the temperatures of the contrastive learning, we use the Hyperopt\footnote{http://hyperopt.github.io/hyperopt/}  hyperparameter selection method to search for the optimal parameters under our framework.

\subsection{Baseline}
In this section, we introduce the baselines we used. Since most of the current research on multilingual fairness focuses on building corpora, there is a lack of methods for debiasing multilingual datasets. The monolingual fairness model relies more on external resources such as sensitive attribute dictionaries, and lacks scalability. Therefore, in the selection of comparison methods, we only selected the following five methods:

• BERT (Individual Training): Since our debiasing method is applied to BERT, we fine-tune the BERT model to measure performance and fairness on the hate speech identification task as a comparison. Similar to \cite{huang-etal-2020-multilingual}, we used each language's dataset to separately train the BERT-based hate speech classification model, which we called it "individual training".

• BERT (Merge Training): Different from the individual training mode, the merge training mode utilizes all of the languages in the multilingual dataset and only trains one hate speech classification model.

• BERT+FGM: One of the most common methods for model debiasing was adversarial learning \citep{chang2020adversarial}. We design an adversarial learning framework that fits our task. We follow the adversarial training method of FGM \citep{miyato2016adversarial} to add perturbation to BERT’s representation module. The disturbance prevents the model from accurately identifying the author’s attributes.

• BERT+PGD: We use another adversarial learning method PGD \citep{madry2017towards}. FGM directly calculates the adversarial disturbance through the “epsilon” parameter at once, so the obtained disturbance may not be optimal. While PGD finds the optimal perturbation through multiple iterations.

• SENT-DEBIAS BERT: This is the sentence-level debiasing method proposed by \cite{liang2020towards}.

\subsection{Main Experiments Results}

\begin{table}
\label{tab:accents2}
\caption{Experimental results of different models on age attribute debiasing}
\centering
\begin{tabular}{c|c|cccc|cccc}
\hline
\multirow{2}{*}{Method} & \multirow{2}{*}{Language} & \multicolumn{4}{c}{Performance ($\uparrow$)} & \multicolumn{4}{|c}{Fairness ($\downarrow$)} \\
& & Acc& $F_m$& $F_w$& AUC & MED & MUED & MEPD & SD \\
\hline
\multirow{6}{*}{\makecell{BERT \\ (Individual \\ Training)}} & English & 89.18 & 86.30 & 89.13 & 0.9324 & 0.0982 & \multirow{6}{*}{-} & \multirow{6}{*}{-} & \multirow{6}{*}{-} \\
 & Italian & 69.20 & 63.76 & 70.10 & 0.7268 & 0.0225 &  &  &  \\
 & Spanish & 69.96 & 69.54 & 70.33 & 0.7852 & 0.0512 &  &  & \\
 & Polish & 86.45 & 65.02 & 84.61 & 0.7875 & 0.0586 &  &  & \\
 & Portuguese & 68.42 & 65.49 & 67.65 & 0.6793 & 0.2292 &  &  & \\ 
 & Avg & 76.64 & 70.02 & 76.36 & 0.7822 & 0.0919 &  &  & \\
\hline
\multirow{6}{*}{\makecell{BERT \\ (Merge \\ Training)}} & English & 88.22 & 85.13 & 88.18 & 0.9247 & 0.0583 & \multirow{6}{*}{0.0306} & \multirow{6}{*}{0.0811} & \multirow{6}{*}{-} \\
 & Italian & 72.43 & 65.17 & 72.34 & 0.7224 & 0.0554 &  &  & \\
 & Spanish & 71.85 & 71.58 & 72.20 & 0.8016 & 0.0284 &  &  & \\
 & Polish & 87.48 & 64.40 & 84.91 & 0.7431 & 0.0012 &  &  & \\
 & Portuguese & 61.40 & 54.79 & 58.49 & 0.5996 & 0.1792 &  &  & \\ 
 & Avg & 76.28 & 68.21 & 75.22 & 0.7583 & 0.0645 &  &  & \\ 
\hline
\multirow{6}{*}{FGM} & English & 88.90 & 85.99 & 88.87 & 0.9318 & 0.0885 & \multirow{6}{*}{0.0472} & \multirow{6}{*}{0.0709} & \multirow{6}{*}{0.0383} \\ 
 & Italian & 71.16 & 64.79 & 71.55 & 0.7520 & 0.0530 &  &  & \\
 & Spanish & 64.55 & 64.53 & 64.35 & 0.7952 & 0.1221 &  &  & \\
 & Polish & 87.91 & 63.45 & 84.85 & 0.8677 & 0.0051 &  &  & \\
 & Portuguese & 65.61 & 62.59 & 64.87 & 0.6862 & 0.0125 &  &  & \\ 
 & Avg & 75.63 & 68.27 & 74.89 & 0.8066  & 0.0562  &  &  & \\
\hline
\multirow{6}{*}{PGD} & English & 87.35 & 84.66 & 87.56 & 0.9234 & 0.0897 & \multirow{6}{*}{0.0358} & \multirow{6}{*}{0.0647} & \multirow{6}{*}{0.0255} \\
 & Italian & 68.51 & 64.16 & 69.79 & 0.7241 & 0.1529 &  &  & \\
 & Spanish & 70.23 & 69.66 & 70.58 & 0.7875 & 0.0509 &  &  & \\
 & Polish & 87.67 & 63.59 & 84.78 & 0.8541 & 0.0112 &  &  & \\
 & Portuguese & 67.02 & 63.27 & 65.78 & 0.7099 & 0.3417 &  &  & \\ 
 & Avg & 76.16 & 69.07 & 75.70  & 0.7998 & 0.1293 &  &  & \\
\hline
\multirow{6}{*}{SentBias} & English & 89.84 & 87.23 & 89.83 & 0.8983  & 0.1128  & \multirow{6}{*}{0.0774} & \multirow{6}{*}{0.0997} & \multirow{6}{*}{0.0007} \\
 & Italian & 70.24 & 49.33 & 64.01 & 0.4970 & 0.0201 &  &  & \\
 & Spanish & 60.22 & 39.42 & 47.25 & 0.4872 & 0.0063 &  &  & \\
 & Polish & 58.25 & 37.57 & 45.26 & 0.3512 & 0.0201 &  &  & \\
 & Portuguese & 58.95 & 40.73 & 47.76 & 0.3860 & 0.0204 &  &  & \\ 
 & Avg & 67.50 & 50.86 & 58.82 & 0.5239 & 0.0359 &  &  & \\
\hline
\multirow{6}{*}{Our Method} & English & 88.69  & 85.93 & 88.74 & 0.9271 & 0.0964 & \multirow{6}{*}{0.0159} & \multirow{6}{*}{0.0736} & \multirow{6}{*}{0.0235} \\
 & Italian & 70.93 & 63.22 & 70.82 & 0.7248  & 0.0328  &  &  & \\
 & Spanish & 71.18 & 70.74 & 71.53 & 0.8050  & 0.0123  &  &  & \\
 & Polish & 87.30 & 68.01 & 85.79 & 0.8068  & 0.0146  &  &  & \\
 & Portuguese & 63.51 & 57.78 & 61.11 & 0.6438  & 0.1120  &  &  & \\ 
 & Avg & 76.32 & 69.14 & 75.60 & 0.7815  & 0.0536  &  &  & \\
\hline
\end{tabular}
\end{table}

\begin{table}
\label{tab:accents3}
\caption{Experimental results of different models on gender attribute debiasing}
\centering
\begin{tabular}{c|c|cccc|cccc}
\hline
\multirow{2}{*}{Method} & \multirow{2}{*}{Language} & \multicolumn{4}{c}{Performance ($\uparrow$)} & \multicolumn{4}{|c}{Fairness ($\downarrow$)} \\
& & Acc& $F_m$& $F_w$& AUC & MED & MUED & MEPD & SD \\
\hline

\multirow{6}{*}{\makecell{BERT \\ (Individual \\ Training)}} & English & 89.18 & 86.30 & 89.13 & 0.9324 & 0.0496 & \multirow{6}{*}{-} & \multirow{6}{*}{-} & \multirow{6}{*}{-} \\
 & Italian & 69.20 & 63.76 & 70.10 & 0.7268 & 0.1405 &  &  & \\
 & Spanish & 69.96 & 69.54 & 70.33 & 0.7852 & 0.1795 &  &  & \\
 & Polish & 86.45 & 65.02 & 84.61 & 0.7875 & 0.0416 &  &  & \\
 & Portuguese & 68.42 & 65.49 & 67.65 & 0.6793 & 0.1458 &  &  & \\
 & Avg & 76.64 & 70.02 & 76.36 & 0.7822 & 0.1114 &  &  & \\
\hline
\multirow{6}{*}{\makecell{BERT \\ (Merge \\ Training)}} & English & 88.22 & 85.13 & 88.18 & 0.9247 & 0.0583 & \multirow{6}{*}{0.0126} & \multirow{6}{*}{0.0811} & \multirow{6}{*}{-} \\
 & Italian & 72.43 & 65.17 & 72.34 & 0.7224 & 0.0554 &  &  & \\
 & Spanish & 71.85 & 71.58 & 72.20 & 0.8016 & 0.0284 &  &  & \\
 & Polish & 87.48 & 64.40 & 84.91 & 0.7431 & 0.0012 &  &  & \\
 & Portuguese & 61.40 & 54.79 & 58.49 & 0.5996 & 0.1792 &  &  & \\ 
 & Avg & 76.28 & 68.21 & 75.22 & 0.7583 & 0.0645 &  &  & \\
 \hline
 \multirow{6}{*}{FGM} & English & 88.90 & 85.99 & 88.87 & 0.9318 & 0.0588 & \multirow{6}{*}{0.0104} & \multirow{6}{*}{0.0709} & \multirow{6}{*}{0.0370} \\
 & Italian & 71.16 & 64.79 & 71.55 & 0.7520 & 0.0798 &  &  & \\
 & Spanish & 64.55 & 64.53 & 64.35 & 0.7952 & 0.1760 &  &  & \\
 & Polish & 87.91 & 63.45 & 84.85 & 0.8677 & 0.0151 &  &  & \\
 & Portuguese & 65.61 & 62.59 & 64.87 & 0.6862 & 0.0125 &  &  & \\
 & Avg & 75.63 & 68.27 & 74.89 & 0.8066  & 0.0685  &  &  & \\
 \hline
 \multirow{6}{*}{PGD} & English & 87.35 & 84.66 & 87.56 & 0.9234 & 0.0790 & \multirow{6}{*}{0.0088} & \multirow{6}{*}{0.0647} & \multirow{6}{*}{0.0440} \\
 & Italian & 68.51 & 64.16 & 69.79 & 0.7241 & 0.0761 &  &  & \\
 & Spanish & 70.23 & 69.66 & 70.58 & 0.7875 & 0.0924 &  &  & \\
 & Polish & 87.67 & 63.59 & 84.78 & 0.8541 & 0.0133 &  &  & \\
 & Portuguese & 67.02 & 63.27 & 65.78 & 0.7099 & 0.1208 &  &  & \\ 
 & Avg & 76.16 & 69.07 & 75.70  & 0.7998 & 0.0763  &  &  & \\
\hline
\multirow{6}{*}{SentBias} & English & 89.84 & 87.23 & 89.83 & 0.8983  & 0.0557 & \multirow{6}{*}{0.0774} & \multirow{6}{*}{0.0997} & \multirow{6}{*}{0.0007} \\
 & Italian & 70.24 & 49.33 & 64.01 & 0.4970 & 0.0354 &  &  & \\
 & Spanish & 60.22 & 39.42 & 47.25 & 0.4872 & 0.0102 &  &  & \\
 & Polish & 58.25 & 37.57 & 45.26 & 0.3512 & 0.0208 &  &  & \\
 & Portuguese & 58.95 & 40.73 & 47.76 & 0.3860 & 0.0208 &  &  & \\ 
 & Avg & 67.50 & 50.86 & 58.82 & 0.5239 & 0.0286 &  &  & \\
\hline
\multirow{6}{*}{Our Method} & English & 88.12  & 84.71  & 87.96  & 0.9283  & 0.0622  & \multirow{6}{*}{0.0028} & \multirow{6}{*}{0.0653} & \multirow{6}{*}{0.0197} \\
 & Italian & 69.32  & 62.53  & 69.72  & 0.7104  & 0.0341  &  &  & \\
 & Spanish & 70.37  & 70.14  & 70.71  & 0.8041  & 0.0211  &  &  & \\
 & Polish & 85.35  & 67.23  & 84.66  & 0.8110  & 0.0409  &  &  & \\ 
 & Portuguese & 65.61  & 61.71  & 64.33  & 0.6281  & 0.0208  &  &  & \\
 & Avg & 75.75  & 69.26  & 75.48  & 0.7764  & 0.0358  &  &  & \\
\hline
\end{tabular}
\end{table}

\begin{table}
\label{tab:accents4}
\caption{Experimental results of different models on ethnicity attribute debiasing}
\centering
\begin{tabular}{c|c|cccc|cccc}
\hline
\multirow{2}{*}{Method} & \multirow{2}{*}{Language} & \multicolumn{4}{c}{Performance ($\uparrow$)} & \multicolumn{4}{|c}{Fairness ($\downarrow$)} \\
& & Acc& $F_m$& $F_w$& AUC & MED & MUED & MEPD & SD \\
\hline
\multirow{6}{*}{\makecell{BERT \\ (Individual \\ Training)}} & English & 89.18 & 86.30 & 89.13 & 0.9324 & 0.0493 & \multirow{6}{*}{-} & \multirow{6}{*}{-} & \multirow{6}{*}{-} \\
 & Italian & 69.20 & 63.76 & 70.10 & 0.7268 & - &  &  & \\ 
 & Spanish & 69.96 & 69.54 & 70.33 & 0.7852 & 0.0610 &  &  & \\ 
 & Polish & 86.45 & 65.02 & 84.61 & 0.7875 & - &  &  & \\
 & Portuguese & 68.42 & 65.49 & 67.65 & 0.6793 & 0.0083 &  &  & \\ 
 & Avg & 76.64 & 70.02 & 76.36 & 0.7822 & 0.0395 &  &  & \\
 \hline
 \multirow{6}{*}{\makecell{BERT \\ (Merge \\ Training)}} & English & 88.22 & 85.13 & 88.18 & 0.9247 & 0.0562 & \multirow{6}{*}{0.0540} & \multirow{6}{*}{0.0811} & \multirow{6}{*}{-} \\
 & Italian & 72.43 & 65.17 & 72.34 & 0.7224 & - &  &  & \\
 & Spanish & 71.85 & 71.58 & 72.20 & 0.8016 & 0.0231 &  &  & \\
 & Polish & 87.48 & 64.40 & 84.91 & 0.7431 & - &  &  & \\
 & Portuguese & 61.40 & 54.79 & 58.49 & 0.5996 & 0.0042 &  &  & \\
 & Avg & 76.28 & 68.21 & 75.22 & 0.7583 & 0.0278 &  &  & \\
 \hline
 \multirow{6}{*}{FGM} & English & 88.90 & 85.99 & 88.87 & 0.9318 & 0.0610 & \multirow{6}{*}{0.0589} & \multirow{6}{*}{0.0709} & \multirow{6}{*}{0.0181} \\
 & Italian & 71.16 & 64.79 & 71.55 & 0.7520 & - &  &  & \\
 & Spanish & 64.55 & 64.53 & 64.35 & 0.7952 & 0.1005 &  &  & \\
 & Polish & 87.91 & 63.45 & 84.85 & 0.8677 & - &  &  & \\
 & Portuguese & 65.61 & 62.59 & 64.87 & 0.6862 & 0.1042 &  &  & \\
 & Avg & 75.63 & 68.27 & 74.89 & 0.8066  & 0.0886 &  &  & \\
 \hline
 \multirow{6}{*}{PGD} & English & 87.35 & 84.66 & 87.56 & 0.9234 & 0.0600  & \multirow{6}{*}{0.0340} & \multirow{6}{*}{0.0647} & \multirow{6}{*}{0.0431} \\
 & Italian & 68.51 & 64.16 & 69.79 & 0.7241 & - &  &  & \\
 & Spanish & 70.23 & 69.66 & 70.58 & 0.7875 & 0.0511  &  &  & \\
 & Polish & 87.67 & 63.59 & 84.78 & 0.8541 & - &  &  & \\
 & Portuguese & 67.02 & 63.27 & 65.78 & 0.7099 & 0.0167  &  &  & \\
 & Avg & 76.16 & 69.07 & 75.70  & 0.7998 & 0.0426  &  &  & \\
\hline
\multirow{6}{*}{SentBias} & English & 89.84 & 87.23 & 89.83 & 0.8983  & 0.0528  & \multirow{6}{*}{0.0286} & \multirow{6}{*}{0.0997} & \multirow{6}{*}{0.0000} \\
 & Italian & 70.24 & 49.33 & 64.01 & 0.4970 & - &  &  & \\
 & Spanish & 60.22 & 39.42 & 47.25 & 0.4872 & 0.0165  &  &  & \\
 & Polish & 58.25 & 37.57 & 45.26 & 0.3512 & - &  &  & \\
 & Portuguese & 58.95 & 40.73 & 47.76 & 0.3860 & 0.0208  &  &  & \\
 & Avg & 67.50 & 50.86 & 58.82 & 0.5239 & 0.0300  &  &  & \\
 \hline
 \multirow{6}{*}{Our Method} & English & 88.72 & 85.85 & 88.72 & 0.9267 & 0.0428 & \multirow{6}{*}{0.0443} & \multirow{6}{*}{0.0795} & \multirow{6}{*}{0.0050} \\
 & Italian & 71.05 & 61.61 & 70.19 & 0.7334 & - &  &  & \\
 & Spanish & 70.77 & 70.47 & 71.13 & 0.7864 & 0.0085  &  &  & \\
 & Polish & 87.73 & 65.67 & 85.36 & 0.8292 & - &  &  & \\
 & Portuguese & 64.56 & 57.52 & 61.23 & 0.5903 & 0.0167  &  &  & \\ 
 & Avg & 76.57 & 68.22 & 75.33 & 0.7732 & 0.0227  &  &  & \\
 \hline
\end{tabular}
\end{table}

\begin{table}
\label{tab:accents5}
\caption{Experimental results of different models on country attribute debiasing}
\centering
\begin{tabular}{c|c|cccc|cccc}
\hline
\multirow{2}{*}{Method} & \multirow{2}{*}{Language} & \multicolumn{4}{c}{Performance ($\uparrow$)} & \multicolumn{4}{|c}{Fairness ($\downarrow$)} \\
& & Acc& $F_m$& $F_w$& AUC & MED & MUED & MEPD & SD \\
\hline
\multirow{6}{*}{\makecell{BERT \\ (Individual \\ Training)}} & English & 89.18 & 86.30 & 89.13 & 0.9324 & 0.0735 & \multirow{6}{*}{-} & \multirow{6}{*}{-} & \multirow{6}{*}{-} \\
 & Italian & 69.20 & 63.76 & 70.10 & 0.7268 & - &  &  & \\
 & Spanish & 69.96 & 69.54 & 70.33 & 0.7852 & 0.0531 &  &  & \\
 & Polish & 86.45 & 65.02 & 84.61 & 0.7875 & - &  &  & \\
 & Portuguese & 68.42 & 65.49 & 67.65 & 0.6793 & 0.1678 &  &  & \\
 & Avg & 76.64 & 70.02 & 76.36 & 0.7822 & 0.0981 &  &  & \\
\hline
 \multirow{6}{*}{\makecell{BERT \\ (Merge \\ Training)}} & English & 88.22 & 85.13 & 88.18 & 0.9247 & 0.0431  & \multirow{6}{*}{0.0059} & \multirow{6}{*}{0.0811} & \multirow{6}{*}{-} \\
 & Italian & 72.43 & 65.17 & 72.34 & 0.7224 & - &  &  & \\
 & Spanish & 71.85 & 71.58 & 72.20 & 0.8016 & 0.0395  &  &  & \\
 & Polish & 87.48 & 64.40 & 84.91 & 0.7431 & - &  &  & \\
 & Portuguese & 61.40 & 54.79 & 58.49 & 0.5996 & 0.0861  &  &  & \\
 & Avg & 76.28 & 68.21 & 75.22 & 0.7583 & 0.0562  &  &  & \\
 \hline
 \multirow{6}{*}{FGM} & English & 88.90 & 85.99 & 88.87 & 0.9318 & 0.0542 & \multirow{6}{*}{0.0004} & \multirow{6}{*}{0.0709} & \multirow{6}{*}{0.0216} \\
 & Italian & 71.16 & 64.79 & 71.55 & 0.7520 & - &  &  & \\
 & Spanish & 64.55 & 64.53 & 64.35 & 0.7952 & 0.0488 &  &  & \\
 & Polish & 87.91 & 63.45 & 84.85 & 0.8677 & - &  &  & \\
 & Portuguese & 65.61 & 62.59 & 64.87 & 0.6862 & 0.2165 &  &  & \\
 & Avg & 75.63 & 68.27 & 74.89 & 0.8066  & 0.1065 &  &  & \\
 \hline
 \multirow{6}{*}{PGD} & English & 87.35 & 84.66 & 87.56 & 0.9234 & 0.0589 & \multirow{6}{*}{0.0391} & \multirow{6}{*}{0.0647} & \multirow{6}{*}{0.0313} \\
 & Italian & 68.51 & 64.16 & 69.79 & 0.7241 & - &  &  & \\
 & Spanish & 70.23 & 69.66 & 70.58 & 0.7875 & 0.1147 &  &  & \\
 & Polish & 87.67 & 63.59 & 84.78 & 0.8541 & - &  &  & \\
 & Portuguese & 67.02 & 63.27 & 65.78 & 0.7099 & 0.1451 &  &  & \\
 & Avg & 76.16 & 69.07 & 75.70  & 0.7998 & 0.1062 &  &  & \\
 \hline
 \multirow{6}{*}{SentBias} & English & 89.84 & 87.23 & 89.83 & 0.8983  & 0.0583  & \multirow{6}{*}{0.0343} & \multirow{6}{*}{0.0997} & \multirow{6}{*}{0.0007} \\
 & Italian & 70.24 & 49.33 & 64.01 & 0.4970 & - &  &  & \\
 & Spanish & 60.22 & 39.42 & 47.25 & 0.4872 & 0.0011  &  &  & \\
 & Polish & 58.25 & 37.57 & 45.26 & 0.3512 & - &  &  & \\
 & Portuguese & 58.95 & 40.73 & 47.76 & 0.3860 & 0.0204  &  &  & \\
 & Avg & 67.50 & 50.86 & 58.82 & 0.5239 & 0.0266  &  &  & \\
 \hline
 \multirow{6}{*}{Our Method} & English & 87.34 & 83.88 & 87.24 & 0.9178 & 0.0460 & \multirow{6}{*}{0.0152} & \multirow{6}{*}{0.0636} & \multirow{6}{*}{0.0029} \\
 & Italian & 69.09 & 62.61 & 69.63 & 0.7216 & - &  &  & \\
 & Spanish & 65.90 & 65.89 & 66.04 & 0.7831 & 0.0379 &  &  & \\
 & Polish & 87.00 & 67.47 & 85.50 & 0.8028 & - &  &  & \\
 & Portuguese & 64.56 & 60.03 & 62.91 & 0.6134 & 0.0515 &  &  & \\
 & Avg & 74.78 & 67.97 & 74.26 & 0.7677 & 0.0451 &  &  & \\
 \hline
\end{tabular}
\end{table}

As shown in Tables 2 to 5, we conduct experiments on the four sensitive attributes of age, country, ethnicity, and gender, respectively, and evaluate them in five areas: performance, monolingual equality difference, multilingual equality difference, multilingual equality performance difference, strategy destructiveness, where the performance indicators include four evaluation indicators: accuracy, macro-average F-value, micro-evaluation F-value, and AUC. We will analyze the performance of our approach in relation to the comparative approach in these five dimensions. 

\textbf{Monolingual Equality Difference} The results of the debiasing experiments for the four sensitive attributes show that training one model for all languages together can significantly reduce the bias towards sensitive attributes. SentBias works best when debiasing the age, gender and country attributes, our method is second only to the SentBias method for these three attributes, and achieves the best results when debiasing the race attribute. It means that our method has good debiasing capabilities. Although our method is not as effective as the Sentbias method in terms of the debiasing effect of the three attributes, we believe that SentBias sacrifices too much model performance, thus allowing the model to "under-fit" and create a "false fairness", meaning that the model tends to predict the majority of samples as the easy category. The FGM method is effective in debiasing age attributes, while training for other attributes fails to achieve "debiasing". Similarly, PGD performs poorly in terms of monolingual equality difference for all four attributes, suggesting that the two methods, FGM and PGD, are unable to focus well on the fairness of a specific language when the task scenario is set in a multilingual mode, with data from all languages trained together. In summary, our model achieves good performance in terms of monolingual equality difference. 

\textbf{Multilingual Equality Difference} In the multilingual equality difference assessment, we can see that our method works best for multilingual assessment when de-biasing the age and gender attributes, reducing the MUED values of BERT (Merge Training) from 0.0306 and 0.0126 to 0.0159 and 0.0028 respectively. The SentBias method has the best debiasing performance when debiasing racial attributes, but it is worth noting that in the multilingual equality difference assessment, the SentBias method is only effective for racial attributes, and for the other three attributes training instead increases the bias for all three attributes. Our method is not optimal in terms of racial attribute de-biasing, but it achieves some debiasing, reducing the MUED value from 0.0540 to 0.0443. Expect for FGM, the other methods have no significant effect on the debiasing of country attributes. In summary, our model has better results as well as more stable performance in terms of multilingual equality difference. 

\textbf{Multilingual equality performance difference} On the equality performance difference assessment, PGD has good performance on the debiasing of the four sensitive attributes, with the MEPD value decreasing from 0.0811 to 0.0647. Our method is second only to the PGD method, with the MEPD values for the four sensitive attributes of age, country, ethnicity and gender decreasing from 0.0811 to 0.0736, 0.0653, 0.0795 and 0.0636, with the MEPD values for the country attribute slightly exceeding those of the PGD model. As for the FGM approach, while it also improves the equality performance difference of the model, the effect is not as pronounced as our approach and the PGD approach, while SentBias has a negative effect in that it exacerbates the performance unfairness across languages. 

\textbf{Strategy destructiveness} We have recently conducted a destructive evaluation of strategies in which the fairness results of BERT (Merge Training) were used as the evaluation results for models that did not employ the debiasing operation. The overall metric trend is that SentBias is the least destructive across sensitive attributes, and our method is slightly more destructive than the SentBias method, but less destructive than the PGD and FGM methods. 

\textbf{Performance} The performance of models trained individually for each language is significantly better than training all languages together in one model, suggesting that joint training between different languages can have some negative gain on each other. When debiasing the age attribute, FGM is less effective than BERT (Merge Training) on several metrics, with only the AUC metric improving, achieving the highest AUC value of 0.8066, while the PGD method only decreases in accuracy, with significant improvements in the other three metrics. When the gender, country, and race attributes are debiased separately, the general experimental picture is similar to that of the age attribute debiasing experiments, except that the model for gender debiasing is slightly lower than BERT (Merge Training) in terms of accuracy, macro-averaged F-value, and micro-evaluated F-value for race debiasing. Collectively, our debiasing method has less negative impact on performance and can even, to some extent, improve the performance of the model. 

Combining the above five categories of metrics, it can be found that our method is not the best performer in terms of a single metric. FGM has the best performance, while SentBias has the best monolingual equality difference and the least strategy destructiveness. In addition, PGD has the best equality performance difference, and our model has the best multilingual equality difference evaluation. Overally, our method is the most comprehensive in terms of performance, meaning that we ensure not only the performance of the model, but also the monolingual equality difference, multilingual equality difference, the equality performance difference, and strategic destructiveness without the large instability of other methods, such as SentBias, which improves the reduced strategic destructiveness but also significantly affects the model's equality performance difference.

\subsection{Ablation Study}

\begin{table}
\label{tab:accents6}
\caption{Ablation experimental results of different models on gender attribute debiasing}
\centering
\begin{tabular}{c|c|cccc|cccc}
\hline
\multirow{2}{*}{Method} & \multirow{2}{*}{Language} & \multicolumn{4}{c}{Performance ($\uparrow$)} & \multicolumn{4}{|c}{Fairness ($\downarrow$)} \\
& & Acc& $F_m$& $F_w$& AUC & MED & MUED & MEPD & SD \\
\hline
\multirow{6}{*}{Our Method} & English & 88.12  & 84.71  & 87.96  & 0.9283  & 0.0622  & \multirow{6}{*}{0.0028} & \multirow{6}{*}{0.0653} & \multirow{6}{*}{0.0197} \\
 & Italian & 69.32  & 62.53  & 69.72  & 0.7104  & 0.0341  &  &  & \\
 & Spanish & 70.37  & 70.14  & 70.71  & 0.8041  & 0.0211  &  &  & \\
 & Polish & 85.35  & 67.23  & 84.66  & 0.8110  & 0.0409  &  &  & \\ 
 & Portuguese & 65.61  & 61.71  & 64.33  & 0.6281  & 0.0208  &  &  & \\
 & Avg & 75.75  & 69.26  & 75.48  & 0.7764  & 0.0358  &  &  & \\
\hline
\multirow{6}{*}{\makecell{- Text \\ Debiasing \\ Module}} & English & 87.84 & 84.64 & 87.80 & 0.9183 & 0.0301 & \multirow{6}{*}{0.0024} & \multirow{6}{*}{0.0416} & \multirow{6}{*}{0.0218} \\
 & Italian & 70.59 & 64.21 & 71.02 & 0.7273 & 0.0341  &  &  & \\
 & Spanish & 69.15 & 68.98 & 69.48 & 0.7996 & 0.1110 &  &  & \\
 & Polish & 86.69 & 65.31 & 84.80 & 0.8064 & 0.0200 &  &  & \\
 & Portuguese & 61.05 & 56.52 & 59.53 & 0.6247 & 0.1000 &  &  & \\
 & Avg & 75.06 & 67.93 & 74.53 & 0.7753 & 0.0590 &  &  & \\
 \hline
 \multirow{6}{*}{\makecell{- Language \\ Fusion \\ Module}} & English & 86.19 & 83.05 & 86.33 & 0.9121 & 0.0551 & \multirow{6}{*}{0.0009} & \multirow{6}{*}{0.0893} & \multirow{6}{*}{0.0157} \\
 & Italian & 70.13 & 63.65 & 70.57 & 0.7313 & 0.0576 &  &  & \\
 & Spanish & 69.01 & 68.86 & 69.34 & 0.8034 & 0.0600 &  &  & \\
 & Polish & 87.30 & 64.18 & 84.77 & 0.7949 & 0.0164 &  &  & \\
 & Portuguese & 65.26 & 61.22 & 63.90 & 0.6481 & 0.1000 &  &  & \\
 & Avg & 75.58 & 68.19 & 74.98 & 0.7780 & 0.0578 &  &  & \\
 \hline
\end{tabular}
\end{table}

\begin{table}
\label{tab:accents7}
\caption{Ablation experimental results of different models on age attribute debiasing}
\centering
\begin{tabular}{c|c|cccc|cccc}
\hline
\multirow{2}{*}{Method} & \multirow{2}{*}{Language} & \multicolumn{4}{c}{Performance ($\uparrow$)} & \multicolumn{4}{|c}{Fairness ($\downarrow$)} \\
& & Acc& $F_m$& $F_w$& AUC & MED & MUED & MEPD & SD \\
\hline
\multirow{6}{*}{Our Method} & English & 88.69  & 85.93 & 88.74 & 0.9271 & 0.0964 & \multirow{6}{*}{0.0159} & \multirow{6}{*}{0.0736} & \multirow{6}{*}{0.0235} \\
 & Italian & 70.93 & 63.22 & 70.82 & 0.7248  & 0.0328  &  &  & \\
 & Spanish & 71.18 & 70.74 & 71.53 & 0.8050  & 0.0123  &  &  & \\
 & Polish & 87.30 & 68.01 & 85.79 & 0.8068  & 0.0146  &  &  & \\
 & Portuguese & 63.51 & 57.78 & 61.11 & 0.6438  & 0.1120  &  &  & \\ 
 & Avg & 76.32 & 69.14 & 75.60 & 0.7815  & 0.0536  &  &  & \\
 \hline
 \multirow{6}{*}{\makecell{- Text \\ Debiasing \\ Module}} & English & 8768 & 8448 & 87.65 & 0.9204 & 0.0955 & \multirow{6}{*}{0.0439} & \multirow{6}{*}{0.0986} & \multirow{6}{*}{0.0281} \\
 & Italian & 69.90 & 64.94 & 70.88 & 0.7348 & 0.0467 &  &  & \\
 & Spanish & 67.79 & 67.78 & 67.93 & 0.7960 & 0.0606 &  &  & \\
 & Polish & 87.55 & 65.63 & 85.27 & 0.8085 & 0.0078 &  &  & \\
 & Portuguese & 63.51 & 59.57 & 62.27 & 0.6080 & 0.0607 &  &  & \\ 
 & Avg & 75.28 & 68.48 & 74.80 & 0.7735 & 0.0543 &  &  & \\
 \hline
 \multirow{6}{*}{\makecell{- Language \\ Fusion \\ Module}} & English & 86.25 & 83.50 & 86.53 & 0.9000 & 0.1009 & \multirow{6}{*}{0.0209} & \multirow{6}{*}{0.0800} & \multirow{6}{*}{0.0178} \\
 & Italian & 71.86 & 63.80 & 71.50 & 0.7286 & 0.0296 &  &  & \\
 & Spanish & 71.18 & 70.20 & 71.39 & 0.7712 & 0.1507 &  &  & \\
 & Polish & 87.36 & 60.97 & 83.93 & 0.7692 & 0.0142 &  &  & \\
 & Portuguese & 67.37 & 62.99 & 65.72 & 0.6616 & 0.0455 &  &  & \\
 & Avg & 76.80 & 68.29 & 75.81 & 0.7661 & 0.0682 &  &  & \\
 \hline
\end{tabular}
\end{table}

\begin{table}
\label{tab:accents8}
\caption{Ablation experimental results of different models on country attribute debiasing}
\centering
\begin{tabular}{c|c|cccc|cccc}
\hline
\multirow{2}{*}{Method} & \multirow{2}{*}{Language} & \multicolumn{4}{c}{Performance ($\uparrow$)} & \multicolumn{4}{|c}{Fairness ($\downarrow$)} \\
& & Acc& $F_m$& $F_w$& AUC & MED & MUED & MEPD & SD \\
\hline
 \multirow{6}{*}{Our Method} & English & 87.34 & 83.88 & 87.24 & 0.9178 & 0.0460 & \multirow{6}{*}{0.0152} & \multirow{6}{*}{0.0636} & \multirow{6}{*}{0.0029} \\
 & Italian & 69.09 & 62.61 & 69.63 & 0.7216 & - &  &  & \\
 & Spanish & 65.90 & 65.89 & 66.04 & 0.7831 & 0.0379 &  &  & \\
 & Polish & 87.00 & 67.47 & 85.50 & 0.8028 & - &  &  & \\
 & Portuguese & 64.56 & 60.03 & 62.91 & 0.6134 & 0.0515 &  &  & \\
 & Avg & 74.78 & 67.97 & 74.26 & 0.7677 & 0.0451 &  &  & \\
 \hline
 \multirow{6}{*}{\makecell{- Text \\ Debiasing \\ Module}} & English & 87.08 & 83.49 & 86.96 & 0.9184 & 0.0342 & \multirow{6}{*}{0.0194}  & \multirow{6}{*}{0.0922}  & \multirow{6}{*}{0.0101} \\ 
 & Italian & 71.05 & 65.02 & 71.57 & 0.7283 & - &  &  & \\
 & Spanish & 69.96 & 69.85 & 70.25 & 0.8075 & 0.0275 &  &  & \\
 & Polish & 87..67 & 66.68 & 85.60 & 0.8191 & - &  &  & \\
 & Portuguese & 63.51 & 59.57 & 62.27 & 0.6228 & 0.1813 &  &  & \\
 & Avg & 75.85 & 68.92 & 75.33 & 0.7792 & 0.0810 &  &  & \\
 \hline
 \multirow{6}{*}{\makecell{- Language \\ Fusion \\ Module}} & English & 87.81 & 84.31 & 87.65 & 0.9233 & 0.0557 & \multirow{6}{*}{0.0174}  & \multirow{6}{*}{0.1054}  & \multirow{6}{*}{0.0203} \\ 
 & Italian & 70.47 & 63.22 & 70.59 & 0.7178 & - &  &  & \\
 & Spanish & 67.39 & 67.35 & 67.59 & 0.7970 & 0.0936 &  &  & \\
 & Polish & 87.30 & 63.98 & 84.72 & 0.8157 & - &  &  & \\
 & Portuguese & 58.95 & 55.80 & 58.33 & 0.5927 & 0.2176  &  &  & \\
 & Avg & 74.38 & 66.93 & 73.77 & 0.7693 & 0.1223 &  &  & \\
 \hline
\end{tabular}
\end{table}

\begin{table}
\label{tab:accents9}
\caption{Ablation experimental results of different models on ethnicity attribute debiasing}
\centering
\begin{tabular}{c|c|cccc|cccc}
\hline
\multirow{2}{*}{Method} & \multirow{2}{*}{Language} & \multicolumn{4}{c}{Performance ($\uparrow$)} & \multicolumn{4}{|c}{Fairness ($\downarrow$)} \\
& & Acc& $F_m$& $F_w$& AUC & MED & MUED & MEPD & SD \\
\hline
\multirow{6}{*}{Our Method} & English & 88.72 & 85.85 & 88.72 & 0.9267 & 0.0428 & \multirow{6}{*}{0.0443} & \multirow{6}{*}{0.0795} & \multirow{6}{*}{0.0050} \\
 & Italian & 71.05 & 61.61 & 70.19 & 0.7334 & - &  &  & \\
 & Spanish & 70.77 & 70.47 & 71.13 & 0.7864 & 0.0085  &  &  & \\
 & Polish & 87.73 & 65.67 & 85.36 & 0.8292 & - &  &  & \\
 & Portuguese & 64.56 & 57.52 & 61.23 & 0.5903 & 0.0167  &  &  & \\ 
 & Avg & 76.57 & 68.22 & 75.33 & 0.7732 & 0.0227  &  &  & \\
 \hline
 \multirow{6}{*}{\makecell{- Text \\ Debiasing \\ Module}} & English & 88.16 & 84.74 & 87.99 & 0.9273 & 0.0465 & \multirow{6}{*}{0.0501} & \multirow{6}{*}{0.1122} & \multirow{6}{*}{0.0119} \\
 & Italian & 69.90 & 65.28 & 70.99 & 0.7141 & - &  &  & \\
 & Spanish & 62.92 & 62.90 & 62.72 & 0.7790 & 0.0338 &  &  & \\
 & Polish & 87.12 & 64.73 & 84.84 & 0.7943 & - &  &  & \\
 & Portuguese & 60.00 & 57.02 & 59.44 & 0.5816 & 0.0750 &  &  & \\
 & Avg & 73.62 & 66.94 & 73.20 & 0.7593 & 0.0518 &  &  & \\
 \hline
 \multirow{6}{*}{\makecell{- Language \\ Fusion \\ Module}} & English & 88.33 & 85.06 & 88.21 & 0.9236 & 0.0548 & \multirow{6}{*}{0.0476} & \multirow{6}{*}{0.1092} & \multirow{6}{*}{0.0355} \\
 & Italian & 68.17 & 63.67 & 69.43 & 0.7324 & - &  &  & \\
 & Spanish & 69.28 & 69.05 & 69.64 & 0.7898 & 0.0728 &  &  & \\
 & Polish & 87.18 & 63.84 & 84.63  & 0.8067 & - &  &  & \\
 & Portuguese & 57.54 & 54.65 & 57.10 & 0.5734 & 0.0208 &  &  & \\
 & Avg & 74.10 & 67.25 & 73.80 & 0.7652 & 0.0495 &  &  & \\
 \hline
\end{tabular}
\end{table}
We further conduct ablation studies on the text debiasing module and the language fusion module, and the results are shown in Tables 6 - 9. We will analyze the results of the debiasing experiments for our approach in five dimensions. 

\textbf{Performance} After removing the text debiasing module, the classification performance of the model largely degrades, except for the negative gain from the module shown in the results of the country attribute debiasing experiments in Table 8. Overall, the text debiasing module has a beneficial effect on the performance of the model itself for the classification task. With the removal of the language fusion module, most of the classification performance evaluation indicators show a significant decrease. 

\textbf{Monolingual equality difference} On the four sensitive attribute debiasing models, after removing the text debiasing module or the language fusion module, the monolingual equality difference of the models increased, meaning that more bias was generated, indicating that these two models had a significant effect in improving the monolingual equality difference of the models. 

\textbf{Multilingual equality difference} Except for the gender attribute debiasing experiment in Table 6, where the removal of the language fusion module reduces the model's multilingual bias, all other ablation experiments demonstrate the effectiveness of text debiasing module and language fusion module in reducing multilingual equality difference under multilingual fairness assessment. 

\textbf{Multilingual equality performance difference} Except for the gender attribute debiasing experiments in Table 6, where the performance fairness of the model across languages is improved after removing the text debiasing module, all other ablation experiments demonstrate the effectiveness of the text debiasing module and the language fusion module in reducing multilingual performance bias. 

\textbf{Strategy destructiveness} The use of the language fusion module is effective in reducing the damage of the debiasing strategy on other sensitive attributes in each of the sensitive attribute debiasing model experiments, whereas the strategy destructiveness of the text debiasing module is relatively unstable and does not increase the bias on other sensitive attributes when debiasing on country or ethnicity, while it significantly increases the bias on other sensitive attributes when debiasing on gender or age.

In summary, after removing the language fusion module, the language performance fairness and the performance of the model are reduced. In addition, in the debiasing experiments of most sensitive attributes, removing the language fusion module will increase the multilingual equality difference of the model and the damage of the debiasing strategy for other sensitive attributes. After removing the debiasing module, the model's performance, monolingual bias, multilingual bias and multilingual performance bias have all rose, but in the damage of the debiasing strategy for other sensitive attributes, there has been an unstable phenomenon, that is, the use of this module may increase Damage to other bias attributes. Experimental results demonstrate that two modules have a significant effect on dealing with multilingual text classification fairness tasks. 

\section{Conclusion}
In this paper, we focus on the task of multilingual text classification and propose a debiasing framework for multilingual text classification based on contrastive learning. Our proposed method does not need to rely on any external language resources and can be extended to any other languages. What's more, We not only focus on the performance ability of the model on the multilingual fairness task, but also hope to evaluate the fairness of the multilingual classification model from more dimensions. So we propose a multi-dimensional fairness evaluation framework for multilingual text classification. 

We hope that our work can provide a more general debiasing method and a more comprehensive evaluation framework for multilingual text fairness tasks. Experimental results demonstrate the effectiveness of our method. Although our method is not the best in a single metric, but overall, our method is the most comprehensive, that is, we not only guarantee the performance of the model, but also guarantee the model's monolingual equality difference, multilingual equality difference, multilingual equality performance difference and strategy destructiveness. Our research can be used as a strong benchmark and evaluation method for multilingual text classification.

\acks{This work was supported by the Ministry of education of Humanities and Social Science project (No. 19YJAZH128 and No. 20YJAZH118) and the Science and Technology Plan Project of Guangzhou (No. 202102080305).}

\vskip 0.2in
\bibliography{sample}

\begin{thebibliography}{43}
\providecommand{\natexlab}[1]{#1}
\providecommand{\url}[1]{\texttt{#1}}
\expandafter\ifx\csname urlstyle\endcsname\relax
  \providecommand{\doi}[1]{doi: #1}\else
  \providecommand{\doi}{doi: \begingroup \urlstyle{rm}\Url}\fi

\bibitem[Bergstra et~al.(2011)Bergstra, Bardenet, Bengio, and
  K{\'e}gl]{bergstra2011algorithms}
James Bergstra, R{\'e}mi Bardenet, Yoshua Bengio, and Bal{\'a}zs K{\'e}gl.
\newblock Algorithms for hyper-parameter optimization.
\newblock \emph{Advances in neural information processing systems}, 24, 2011.

\bibitem[Bolukbasi et~al.(2016)Bolukbasi, Chang, Zou, Saligrama, and
  Kalai]{10.5555/3157382.3157584}
Tolga Bolukbasi, Kai-Wei Chang, James Zou, Venkatesh Saligrama, and Adam Kalai.
\newblock Man is to computer programmer as woman is to homemaker? debiasing
  word embeddings.
\newblock In \emph{Proceedings of the 30th International Conference on Neural
  Information Processing Systems}, NIPS'16, page 4356–4364, Red Hook, NY,
  USA, 2016. Curran Associates Inc.
\newblock ISBN 9781510838819.

\bibitem[C{\^a}mara et~al.(2022)C{\^a}mara, Taneja, Azad, Allaway, and
  Zemel]{camara-etal-2022-mapping}
Ant{\'o}nio C{\^a}mara, Nina Taneja, Tamjeed Azad, Emily Allaway, and Richard
  Zemel.
\newblock Mapping the multilingual margins: Intersectional biases of sentiment
  analysis systems in {E}nglish, {S}panish, and {A}rabic.
\newblock In \emph{Proceedings of the Second Workshop on Language Technology
  for Equality, Diversity and Inclusion}, pages 90--106, Dublin, Ireland, May
  2022. Association for Computational Linguistics.
\newblock \doi{10.18653/v1/2022.ltedi-1.11}.
\newblock URL \url{https://aclanthology.org/2022.ltedi-1.11}.

\bibitem[Chang et~al.(2020)Chang, Nguyen, Murakonda, Kazemi, and
  Shokri]{chang2020adversarial}
Hongyan Chang, Ta~Duy Nguyen, Sasi~Kumar Murakonda, Ehsan Kazemi, and Reza
  Shokri.
\newblock On adversarial bias and the robustness of fair machine learning.
\newblock \emph{arXiv preprint arXiv:2006.08669}, 2020.

\bibitem[Cheng et~al.(2021)Cheng, Mosallanezhad, Silva, Hall, and
  Liu]{cheng-etal-2021-mitigating}
Lu~Cheng, Ahmadreza Mosallanezhad, Yasin Silva, Deborah Hall, and Huan Liu.
\newblock Mitigating bias in session-based cyberbullying detection: A
  non-compromising approach.
\newblock In \emph{Proceedings of the 59th Annual Meeting of the Association
  for Computational Linguistics and the 11th International Joint Conference on
  Natural Language Processing (Volume 1: Long Papers)}, pages 2158--2168,
  Online, August 2021. Association for Computational Linguistics.
\newblock \doi{10.18653/v1/2021.acl-long.168}.
\newblock URL \url{https://aclanthology.org/2021.acl-long.168}.

\bibitem[Cheng et~al.(2020)Cheng, Hao, Yuan, Si, and Carin]{cheng2020fairfil}
Pengyu Cheng, Weituo Hao, Siyang Yuan, Shijing Si, and Lawrence Carin.
\newblock Fairfil: Contrastive neural debiasing method for pretrained text
  encoders.
\newblock In \emph{International Conference on Learning Representations}, 2020.

\bibitem[Cho et~al.(2021)Cho, Kim, Yang, and Kim]{10.1145/3442188.3445907}
Won~Ik Cho, Jiwon Kim, Jaeyeong Yang, and Nam~Soo Kim.
\newblock Towards cross-lingual generalization of translation gender bias.
\newblock In \emph{Proceedings of the 2021 ACM Conference on Fairness,
  Accountability, and Transparency}, FAccT '21, page 449–457, New York, NY,
  USA, 2021. Association for Computing Machinery.
\newblock ISBN 9781450383097.
\newblock \doi{10.1145/3442188.3445907}.
\newblock URL \url{https://doi.org/10.1145/3442188.3445907}.

\bibitem[Costa-jussà et~al.(2022)Costa-jussà, Escolano, Basta, Ferrando,
  Batlle, and
  Kharitonova]{Costa-juss_Escolano_Basta_Ferrando_Batlle_Kharitonova_2022}
Marta~R. Costa-jussà, Carlos Escolano, Christine Basta, Javier Ferrando, Roser
  Batlle, and Ksenia Kharitonova.
\newblock Interpreting gender bias in neural machine translation: Multilingual
  architecture matters.
\newblock \emph{Proceedings of the AAAI Conference on Artificial Intelligence},
  36\penalty0 (11):\penalty0 11855--11863, Jun. 2022.
\newblock \doi{10.1609/aaai.v36i11.21442}.
\newblock URL \url{https://ojs.aaai.org/index.php/AAAI/article/view/21442}.

\bibitem[Devlin et~al.(2019)Devlin, Chang, Lee, and
  Toutanova]{devlin-etal-2019-bert}
Jacob Devlin, Ming-Wei Chang, Kenton Lee, and Kristina Toutanova.
\newblock {BERT}: Pre-training of deep bidirectional transformers for language
  understanding.
\newblock In \emph{Proceedings of the 2019 Conference of the North {A}merican
  Chapter of the Association for Computational Linguistics: Human Language
  Technologies, Volume 1 (Long and Short Papers)}, pages 4171--4186,
  Minneapolis, Minnesota, June 2019. Association for Computational Linguistics.
\newblock \doi{10.18653/v1/N19-1423}.
\newblock URL \url{https://aclanthology.org/N19-1423}.

\bibitem[Dixon et~al.(2018)Dixon, Li, Sorensen, Thain, and
  Vasserman]{10.1145/3278721.3278729}
Lucas Dixon, John Li, Jeffrey Sorensen, Nithum Thain, and Lucy Vasserman.
\newblock Measuring and mitigating unintended bias in text classification.
\newblock AIES '18, page 67–73, New York, NY, USA, 2018. Association for
  Computing Machinery.
\newblock ISBN 9781450360128.
\newblock \doi{10.1145/3278721.3278729}.
\newblock URL \url{https://doi.org/10.1145/3278721.3278729}.

\bibitem[Dwork et~al.(2012)Dwork, Hardt, Pitassi, Reingold, and
  Zemel]{10.1145/2090236.2090255}
Cynthia Dwork, Moritz Hardt, Toniann Pitassi, Omer Reingold, and Richard Zemel.
\newblock Fairness through awareness.
\newblock In \emph{Proceedings of the 3rd Innovations in Theoretical Computer
  Science Conference}, ITCS '12, page 214–226, New York, NY, USA, 2012.
  Association for Computing Machinery.
\newblock ISBN 9781450311151.
\newblock \doi{10.1145/2090236.2090255}.
\newblock URL \url{https://doi.org/10.1145/2090236.2090255}.

\bibitem[Elsafoury(2022)]{elsafoury-2022-darkness}
Fatma Elsafoury.
\newblock Darkness can not drive out darkness: Investigating bias in hate
  {S}peech{D}etection models.
\newblock In \emph{Proceedings of the 60th Annual Meeting of the Association
  for Computational Linguistics: Student Research Workshop}, pages 31--43,
  Dublin, Ireland, May 2022. Association for Computational Linguistics.
\newblock URL \url{https://aclanthology.org/2022.acl-srw.4}.

\bibitem[Garc{\'i}a-D{\'i}az et~al.(2022)Garc{\'i}a-D{\'i}az,
  Jim{\'e}nez-Zafra, Garc{\'i}a-Cumbreras, and
  Valencia-Garc{\'i}a]{GarcaDaz2022EvaluatingFC}
Jos{\'e}~Antonio Garc{\'i}a-D{\'i}az, Salud~Mar{\'i}a Jim{\'e}nez-Zafra,
  Miguel~Angel Garc{\'i}a-Cumbreras, and Rafael Valencia-Garc{\'i}a.
\newblock Evaluating feature combination strategies for hate-speech detection
  in spanish using linguistic features and transformers.
\newblock \emph{Complex \& Intelligent Systems}, 2022.

\bibitem[Gupta et~al.(2021)Gupta, Ferber, Dilkina, and Ver~Steeg]{Gupta}
Umang Gupta, Aaron~M Ferber, Bistra Dilkina, and Greg Ver~Steeg.
\newblock Controllable guarantees for fair outcomes via contrastive information
  estimation.
\newblock \emph{Proceedings of the AAAI Conference on Artificial Intelligence},
  35\penalty0 (9):\penalty0 7610--7619, May 2021.
\newblock \doi{10.1609/aaai.v35i9.16931}.
\newblock URL \url{https://ojs.aaai.org/index.php/AAAI/article/view/16931}.

\bibitem[Han et~al.(2021)Han, Baldwin, and Cohn]{han-etal-2021-decoupling}
Xudong Han, Timothy Baldwin, and Trevor Cohn.
\newblock Decoupling adversarial training for fair {NLP}.
\newblock In \emph{Findings of the Association for Computational Linguistics:
  ACL-IJCNLP 2021}, pages 471--477, Online, August 2021. Association for
  Computational Linguistics.
\newblock \doi{10.18653/v1/2021.findings-acl.41}.
\newblock URL \url{https://aclanthology.org/2021.findings-acl.41}.

\bibitem[Huang et~al.(2020)Huang, Xing, Dernoncourt, and
  Paul]{huang-etal-2020-multilingual}
Xiaolei Huang, Linzi Xing, Franck Dernoncourt, and Michael~J. Paul.
\newblock Multilingual twitter corpus and baselines for evaluating demographic
  bias in hate speech recognition.
\newblock In \emph{Proceedings of the Twelfth Language Resources and Evaluation
  Conference}, pages 1440--1448, Marseille, France, May 2020. European Language
  Resources Association.
\newblock ISBN 979-10-95546-34-4.
\newblock URL \url{https://aclanthology.org/2020.lrec-1.180}.

\bibitem[Kiritchenko and Mohammad(2018)]{kiritchenko-mohammad-2018-examining}
Svetlana Kiritchenko and Saif Mohammad.
\newblock Examining gender and race bias in two hundred sentiment analysis
  systems.
\newblock In \emph{Proceedings of the Seventh Joint Conference on Lexical and
  Computational Semantics}, pages 43--53, New Orleans, Louisiana, June 2018.
  Association for Computational Linguistics.
\newblock \doi{10.18653/v1/S18-2005}.
\newblock URL \url{https://aclanthology.org/S18-2005}.

\bibitem[Kocoń et~al.(2021)Kocoń, Figas, Gruza, Puchalska, Kajdanowicz, and
  Kazienko]{KOCON2021102643}
Jan Kocoń, Alicja Figas, Marcin Gruza, Daria Puchalska, Tomasz Kajdanowicz,
  and Przemysław Kazienko.
\newblock Offensive, aggressive, and hate speech analysis: From data-centric to
  human-centered approach.
\newblock \emph{Information Processing and Management}, 58\penalty0
  (5):\penalty0 102643, 2021.
\newblock ISSN 0306-4573.
\newblock \doi{https://doi.org/10.1016/j.ipm.2021.102643}.
\newblock URL
  \url{https://www.sciencedirect.com/science/article/pii/S0306457321001333}.

\bibitem[Kumar et~al.(2022)Kumar, Gehlot, Mullappilly, and
  Nandakumar]{kumar-etal-2022-mucot}
Gokul~Karthik Kumar, Abhishek Gehlot, Sahal~Shaji Mullappilly, and Karthik
  Nandakumar.
\newblock {M}u{C}o{T}: Multilingual contrastive training for question-answering
  in low-resource languages.
\newblock In \emph{Proceedings of the Second Workshop on Speech and Language
  Technologies for Dravidian Languages}, pages 15--24, Dublin, Ireland, May
  2022. Association for Computational Linguistics.
\newblock \doi{10.18653/v1/2022.dravidianlangtech-1.3}.
\newblock URL \url{https://aclanthology.org/2022.dravidianlangtech-1.3}.

\bibitem[Liang et~al.(2020)Liang, Li, Zheng, Lim, Salakhutdinov, and
  Morency]{liang2020towards}
Paul~Pu Liang, Irene~Mengze Li, Emily Zheng, Yao~Chong Lim, Ruslan
  Salakhutdinov, and Louis-Philippe Morency.
\newblock Towards debiasing sentence representations.
\newblock \emph{arXiv preprint arXiv:2007.08100}, 2020.

\bibitem[Madry et~al.(2017)Madry, Makelov, Schmidt, Tsipras, and
  Vladu]{madry2017towards}
Aleksander Madry, Aleksandar Makelov, Ludwig Schmidt, Dimitris Tsipras, and
  Adrian Vladu.
\newblock Towards deep learning models resistant to adversarial attacks.
\newblock \emph{arXiv preprint arXiv:1706.06083}, 2017.

\bibitem[Miyato et~al.(2016)Miyato, Dai, and Goodfellow]{miyato2016adversarial}
Takeru Miyato, Andrew~M Dai, and Ian Goodfellow.
\newblock Adversarial training methods for semi-supervised text classification.
\newblock \emph{arXiv preprint arXiv:1605.07725}, 2016.

\bibitem[Mutuvi et~al.(2020)Mutuvi, Boros, Doucet, Jatowt, Lejeune, and
  Odeo]{mutuvi-etal-2020-multilingual}
Stephen Mutuvi, Emanuela Boros, Antoine Doucet, Adam Jatowt, Ga{\"e}l Lejeune,
  and Moses Odeo.
\newblock Multilingual epidemiological text classification: A comparative
  study.
\newblock In \emph{Proceedings of the 28th International Conference on
  Computational Linguistics}, pages 6172--6183, Barcelona, Spain (Online),
  December 2020. International Committee on Computational Linguistics.
\newblock \doi{10.18653/v1/2020.coling-main.543}.
\newblock URL \url{https://aclanthology.org/2020.coling-main.543}.

\bibitem[Nooralahzadeh et~al.(2020)Nooralahzadeh, Bekoulis, Bjerva, and
  Augenstein]{nooralahzadeh-etal-2020-zero}
Farhad Nooralahzadeh, Giannis Bekoulis, Johannes Bjerva, and Isabelle
  Augenstein.
\newblock Zero-shot cross-lingual transfer with meta learning.
\newblock In \emph{Proceedings of the 2020 Conference on Empirical Methods in
  Natural Language Processing (EMNLP)}, pages 4547--4562, Online, November
  2020. Association for Computational Linguistics.
\newblock \doi{10.18653/v1/2020.emnlp-main.368}.
\newblock URL \url{https://aclanthology.org/2020.emnlp-main.368}.

\bibitem[Pan et~al.(2021)Pan, Wang, Wu, and Li]{pan-etal-2021-contrastive}
Xiao Pan, Mingxuan Wang, Liwei Wu, and Lei Li.
\newblock Contrastive learning for many-to-many multilingual neural machine
  translation.
\newblock In \emph{Proceedings of the 59th Annual Meeting of the Association
  for Computational Linguistics and the 11th International Joint Conference on
  Natural Language Processing (Volume 1: Long Papers)}, pages 244--258, Online,
  August 2021. Association for Computational Linguistics.
\newblock \doi{10.18653/v1/2021.acl-long.21}.
\newblock URL \url{https://aclanthology.org/2021.acl-long.21}.

\bibitem[Ponti et~al.(2020)Ponti, Glava{\v{s}}, Majewska, Liu, Vuli{\'c}, and
  Korhonen]{ponti-etal-2020-xcopa}
Edoardo~Maria Ponti, Goran Glava{\v{s}}, Olga Majewska, Qianchu Liu, Ivan
  Vuli{\'c}, and Anna Korhonen.
\newblock {XCOPA}: A multilingual dataset for causal commonsense reasoning.
\newblock In \emph{Proceedings of the 2020 Conference on Empirical Methods in
  Natural Language Processing (EMNLP)}, pages 2362--2376, Online, November
  2020. Association for Computational Linguistics.
\newblock \doi{10.18653/v1/2020.emnlp-main.185}.
\newblock URL \url{https://aclanthology.org/2020.emnlp-main.185}.

\bibitem[Prabhakaran et~al.(2019)Prabhakaran, Hutchinson, and
  Mitchell]{prabhakaran-etal-2019-perturbation}
Vinodkumar Prabhakaran, Ben Hutchinson, and Margaret Mitchell.
\newblock Perturbation sensitivity analysis to detect unintended model biases.
\newblock In \emph{Proceedings of the 2019 Conference on Empirical Methods in
  Natural Language Processing and the 9th International Joint Conference on
  Natural Language Processing (EMNLP-IJCNLP)}, pages 5740--5745, Hong Kong,
  China, November 2019. Association for Computational Linguistics.
\newblock \doi{10.18653/v1/D19-1578}.
\newblock URL \url{https://www.aclweb.org/anthology/D19-1578}.

\bibitem[Ptaszynski et~al.(2019)Ptaszynski, Pieciukiewicz, and
  Dybała]{publ152265}
Michal Ptaszynski, Agata Pieciukiewicz, and Paweł Dybała.
\newblock Results of the poleval 2019 shared task 6 : first dataset and open
  shared task for automatic cyberbullying detection in polish twitter.
\newblock In Maciej Ogrodniczuk and Łukasz Kobyliński, editors,
  \emph{Proceedings of the PolEval 2019 Workshop}, pages 89--110. Institute of
  Computer Sciences. Polish Academy of Sciences, Warszawa, 2019.
\newblock ISBN 978-83-63159-28-3.
\newblock URL \url{http://2019.poleval.pl/files/poleval2019.pdf}.

\bibitem[Ruder et~al.(2022)Ruder, Vuli{\'c}, and
  S{\o}gaard]{ruder-etal-2022-square}
Sebastian Ruder, Ivan Vuli{\'c}, and Anders S{\o}gaard.
\newblock Square one bias in {NLP}: Towards a multi-dimensional exploration of
  the research manifold.
\newblock In \emph{Findings of the Association for Computational Linguistics:
  ACL 2022}, pages 2340--2354, Dublin, Ireland, May 2022. Association for
  Computational Linguistics.
\newblock \doi{10.18653/v1/2022.findings-acl.184}.
\newblock URL \url{https://aclanthology.org/2022.findings-acl.184}.

\bibitem[Rudinger et~al.(2017)Rudinger, May, and
  Van~Durme]{rudinger-etal-2017-social}
Rachel Rudinger, Chandler May, and Benjamin Van~Durme.
\newblock Social bias in elicited natural language inferences.
\newblock In \emph{Proceedings of the First {ACL} Workshop on Ethics in Natural
  Language Processing}, pages 74--79, Valencia, Spain, April 2017. Association
  for Computational Linguistics.
\newblock \doi{10.18653/v1/W17-1609}.
\newblock URL \url{https://aclanthology.org/W17-1609}.

\bibitem[Shen et~al.(2021)Shen, Han, Cohn, Baldwin, and
  Frermann]{shen2021contrastive}
Aili Shen, Xudong Han, Trevor Cohn, Timothy Baldwin, and Lea Frermann.
\newblock Contrastive learning for fair representations.
\newblock 2021.

\bibitem[Shen et~al.(2018)Shen, Yoder, Jo, and Rose]{Shen_Yoder_Jo_Rose_2018}
Qinlan Shen, Michael Yoder, Yohan Jo, and Carolyn Rose.
\newblock Perceptions of censorship and moderation bias in political debate
  forums.
\newblock \emph{Proceedings of the International AAAI Conference on Web and
  Social Media}, 12\penalty0 (1), Jun. 2018.
\newblock \doi{10.1609/icwsm.v12i1.15002}.
\newblock URL \url{https://ojs.aaai.org/index.php/ICWSM/article/view/15002}.

\bibitem[Wan(2022)]{wan2022fairness}
Ada Wan.
\newblock Fairness in representation for multilingual nlp: Insights from
  controlled experiments on conditional language modeling.
\newblock In \emph{International Conference on Learning Representations}, 2022.
\newblock URL \url{https://openreview.net/forum?id=-llS6TiOew}.

\bibitem[Wang et~al.(2021)Wang, Chen, Zhou, Qiu, and
  Li]{wang-etal-2021-contrastive}
Danqing Wang, Jiaze Chen, Hao Zhou, Xipeng Qiu, and Lei Li.
\newblock Contrastive aligned joint learning for multilingual summarization.
\newblock In \emph{Findings of the Association for Computational Linguistics:
  ACL-IJCNLP 2021}, pages 2739--2750, Online, August 2021. Association for
  Computational Linguistics.
\newblock \doi{10.18653/v1/2021.findings-acl.242}.
\newblock URL \url{https://aclanthology.org/2021.findings-acl.242}.

\bibitem[Wang et~al.(2022)Wang, Liu, and Wang]{wang-etal-2022-assessing}
Jialu Wang, Yang Liu, and Xin Wang.
\newblock Assessing multilingual fairness in pre-trained multimodal
  representations.
\newblock In \emph{Findings of the Association for Computational Linguistics:
  ACL 2022}, pages 2681--2695, Dublin, Ireland, May 2022. Association for
  Computational Linguistics.
\newblock \doi{10.18653/v1/2022.findings-acl.211}.
\newblock URL \url{https://aclanthology.org/2022.findings-acl.211}.

\bibitem[Waseem and Hovy(2016)]{waseem-hovy-2016-hateful}
Zeerak Waseem and Dirk Hovy.
\newblock Hateful symbols or hateful people? predictive features for hate
  speech detection on {T}witter.
\newblock In \emph{Proceedings of the {NAACL} Student Research Workshop}, pages
  88--93, San Diego, California, June 2016. Association for Computational
  Linguistics.
\newblock \doi{10.18653/v1/N16-2013}.
\newblock URL \url{https://aclanthology.org/N16-2013}.

\bibitem[Webster and Pitler(2020)]{DBLP:journals/corr/abs-2006-08881}
Kellie Webster and Emily Pitler.
\newblock Scalable cross lingual pivots to model pronoun gender for
  translation.
\newblock \emph{CoRR}, abs/2006.08881, 2020.
\newblock URL \url{https://arxiv.org/abs/2006.08881}.

\bibitem[Wu and Dredze(2019)]{wu-dredze-2019-beto}
Shijie Wu and Mark Dredze.
\newblock Beto, bentz, becas: The surprising cross-lingual effectiveness of
  {BERT}.
\newblock In \emph{Proceedings of the 2019 Conference on Empirical Methods in
  Natural Language Processing and the 9th International Joint Conference on
  Natural Language Processing (EMNLP-IJCNLP)}, pages 833--844, Hong Kong,
  China, November 2019. Association for Computational Linguistics.
\newblock \doi{10.18653/v1/D19-1077}.
\newblock URL \url{https://aclanthology.org/D19-1077}.

\bibitem[Wu and Dredze(2020)]{wu-dredze-2020-languages}
Shijie Wu and Mark Dredze.
\newblock Are all languages created equal in multilingual {BERT}?
\newblock In \emph{Proceedings of the 5th Workshop on Representation Learning
  for NLP}, pages 120--130, Online, July 2020. Association for Computational
  Linguistics.
\newblock \doi{10.18653/v1/2020.repl4nlp-1.16}.
\newblock URL \url{https://aclanthology.org/2020.repl4nlp-1.16}.

\bibitem[Yin and Zubiaga(2021)]{yin2021towards}
Wenjie Yin and Arkaitz Zubiaga.
\newblock Towards generalisable hate speech detection: a review on obstacles
  and solutions.
\newblock \emph{PeerJ Computer Science}, 7:\penalty0 e598, 2021.

\bibitem[Zhao et~al.(2021)Zhao, Yang, Wang, Yang, and Deng]{ijcai2021p473}
Han Zhao, Xu~Yang, Zhenru Wang, Erkun Yang, and Cheng Deng.
\newblock Graph debiased contrastive learning with joint representation
  clustering.
\newblock In Zhi-Hua Zhou, editor, \emph{Proceedings of the Thirtieth
  International Joint Conference on Artificial Intelligence, {IJCAI-21}}, pages
  3434--3440. International Joint Conferences on Artificial Intelligence
  Organization, 8 2021.
\newblock \doi{10.24963/ijcai.2021/473}.
\newblock URL \url{https://doi.org/10.24963/ijcai.2021/473}.
\newblock Main Track.

\bibitem[Zhao et~al.(2018)Zhao, Wang, Yatskar, Ordonez, and
  Chang]{zhao-etal-2018-gender}
Jieyu Zhao, Tianlu Wang, Mark Yatskar, Vicente Ordonez, and Kai-Wei Chang.
\newblock Gender bias in coreference resolution: Evaluation and debiasing
  methods.
\newblock In \emph{Proceedings of the 2018 Conference of the North {A}merican
  Chapter of the Association for Computational Linguistics: Human Language
  Technologies, Volume 2 (Short Papers)}, pages 15--20, New Orleans, Louisiana,
  June 2018. Association for Computational Linguistics.
\newblock \doi{10.18653/v1/N18-2003}.
\newblock URL \url{https://aclanthology.org/N18-2003}.

\bibitem[Zhao et~al.(2020)Zhao, Mukherjee, Hosseini, Chang, and
  Hassan~Awadallah]{zhao-etal-2020-gender}
Jieyu Zhao, Subhabrata Mukherjee, Saghar Hosseini, Kai-Wei Chang, and Ahmed
  Hassan~Awadallah.
\newblock Gender bias in multilingual embeddings and cross-lingual transfer.
\newblock In \emph{Proceedings of the 58th Annual Meeting of the Association
  for Computational Linguistics}, pages 2896--2907, Online, July 2020.
  Association for Computational Linguistics.
\newblock \doi{10.18653/v1/2020.acl-main.260}.
\newblock URL \url{https://aclanthology.org/2020.acl-main.260}.

\end{thebibliography}

\end{document}